\documentclass[letterpaper]{article} %
\usepackage{aaai2026}  %
\copyrighttext{%
  \phantom{Copyright \copyright\ 2026, Association for the Advancement of Artificial} \\
  \phantom{Intelligence (www.aaai.org). All rights reserved.}%
}
\usepackage{times}  %
\usepackage{helvet}  %
\usepackage{courier}  %
\usepackage[hyphens]{url}  %
\usepackage{graphicx} %
\usepackage{natbib}  %
\usepackage{caption} %
\usepackage{algorithm}
\usepackage{algorithmic}
\usepackage[acronym]{glossaries}
\glsdisablehyper   %

\usepackage{newfloat}
\usepackage{listings}
\DeclareCaptionStyle{ruled}{labelfont=normalfont,labelsep=colon,strut=off} %
\floatstyle{ruled}
\newfloat{listing}{tb}{lst}{}
\floatname{listing}{Listing}
\usepackage{xspace}
\usepackage{paralist}

\usepackage[dvipsnames]{xcolor}%
\usepackage{subcaption}

\newcommand{\eg}{e.\,g.,\xspace}

\newcommand{\cf}{c.\,f.,\xspace}

\newacronym{3dssg}{3DSSG}{3D Semantic Scene Graph}
\newacronym{clip}{CLIP}{Contrastive Language-Image Pretraining}
\newacronym{gnn}{GNN}{Graph Neural Network}
\newacronym{tsdf}{TSDF}{Truncated Signed Distance Function}
\newacronym{vlfm}{VLFM}{Vision-Language Foundation Model}
\newacronym{llm}{LLM}{Large Language Model}
\newacronym{dsg}{DSG}{3D Dynamic Scene Graphs}

\title{A Scene Graph Backed Approach to Open Set Semantic Mapping}

\author{
    Martin Günther\equalcontrib\textsuperscript{\rm 1},
    Felix Igelbrink\equalcontrib\textsuperscript{\rm 1},
    Oscar Lima\equalcontrib\textsuperscript{\rm 1},
    Lennart Niecksch\equalcontrib\textsuperscript{\rm 1, \rm 2},
    Marian Renz\equalcontrib\textsuperscript{\rm 1, \rm 2},\\
    Martin Atzmueller\textsuperscript{\rm 2,\rm 1}
}
\affiliations{
    \textsuperscript{\rm 1}German Research Center for Artificial Intelligence (DFKI),\\ Cooperative and Autonomous Systems (CAS), Hamburger Straße 24, Osnabrück, Germany\\
    \textsuperscript{\rm 2}Osnabrück University, Semantic Information Systems Group, Wachsbleiche 27, Osnabrück, Germany\\
    \{martin.guenther, felix.igelbrink, oscar.lima, lennart.niecksch, marian.renz\}@dfki.de, martin.atzmueller@uos.de
}

\begin{document}

\maketitle

\begin{abstract}
While Open Set Semantic Mapping and 3D Semantic Scene Graphs (3DSSGs) are established paradigms in robotic perception, deploying them effectively to support high-level reasoning in large-scale, real-world environments remains a significant challenge. Most existing approaches decouple perception from representation, treating the scene graph as a derivative layer generated post hoc. This limits both consistency and scalability. In contrast, we propose a mapping architecture where the 3DSSG serves as the foundational backend, acting as the primary knowledge representation for the entire mapping process.

Our approach leverages prior work on incremental scene graph prediction to infer and update the graph structure in real-time as the environment is explored. This ensures that the map remains topologically consistent and computationally efficient, even during extended operations in large-scale settings. By maintaining an explicit, spatially grounded representation that supports both flat and hierarchical topologies, we bridge the gap between sub-symbolic raw sensor data and high-level symbolic reasoning. Consequently, this provides a stable, verifiable structure that knowledge-driven frameworks, ranging from knowledge graphs and ontologies to Large Language Models (LLMs), can directly exploit, enabling agents to operate with enhanced interpretability, trustworthiness, and alignment to human concepts.

\end{abstract}

\begin{links}
    \link{Code and video}{https://dfki-ni.github.io/SSG-MAKE-2026/}
\end{links}

\section{Introduction}
\label{sec:introduction}

Autonomous mobile robots operating in large-scale, real-world environments require a semantic understanding that extends beyond pure geometric mapping to effectively perform complex tasks. While volumetric representations like \glspl{tsdf} are highly effective for integrating geometric information and producing high-fidelity surface reconstructions, they fundamentally lack an explicit model of the scene's structure. By treating the environment as a monolithic volume or a collection of geometric primitives, these approaches often decouple perception from high-level representation, making it difficult to model the hierarchical and semantic relationships necessary for advanced reasoning.

To bridge this gap between sub-symbolic sensor data and symbolic reasoning, \glspl{3dssg} have emerged as a powerful abstraction, representing environments as hierarchical graphs of entities and their relationships. Unlike implicit representations or unstructured point clouds, a \gls{3dssg} acts as an explicit knowledge structure. It achieves this by discretizing the continuous sensor stream into distinct symbolic entities (nodes) and encoding their semantic interactions as edges. This structure aligns the robot's internal representation with human-understandable concepts, providing a stable grounding for higher-level logic. By formally defining these relationships---whether spatial (\eg \emph{on top of}) or semantic (\eg \emph{part of})---the \gls{3dssg} transforms raw perception into a queryable knowledge base. This allows the system to not only store where geometry is located but to reason about what the environment is, enabling the integration of external knowledge priors and facilitating open-set reasoning.

However, many existing frameworks treat the scene graph as a derivative layer generated post-hoc, which limits the system's ability to maintain consistency over extended operations and prevents online interactions. In this work, we instead propose a mapping architecture where the \gls{3dssg} serves as the foundational backend for the entire mapping process. Rather than relying solely on dense reconstruction for all tasks, our presented approach uses the \gls{3dssg} to maintain an explicit, spatially grounded representation that supports both flat and hierarchical topologies. This structure allows knowledge-driven frameworks, ranging from knowledge graphs to \glspl{llm}, to directly exploit the structure of the scene, enabling agents to operate with enhanced interpretability and alignment to human concepts.

In this paper, we present the following contributions:
\begin{itemize}
    \item We describe a modular mapping architecture that establishes the \gls{3dssg} as the foundational backend and primary knowledge representation (Section~\ref{sec:mapping-approach}). This design moves beyond rigid hierarchies to support dynamic, modular layers that efficiently separate spatial from semantic relations, utilizing voxel-based representations for scalability.
    \item We introduce a method for inferring and updating the graph structure in real-time as the environment is explored (Sections~\ref{sec:mapping-approach} and \ref{sec:ipp}). This ensures topological consistency by integrating bottom-up sensory data with top-down expectations derived from prior observations.
    \item We demonstrate the versatility of our framework through three key integrations (Section~\ref{sec:results}):
    \begin{itemize}
        \item \emph{Dynamic Open-Vocabulary Queries:} Incorporating \glspl{vlfm} such as \gls{clip} to facilitate natural language interaction.
        \item \emph{Real-Time Relation Prediction:} Integrating a \gls{3dssg} prediction network to infer structural relationships dynamically during operation.
        \item \emph{Real-World Deployment:} Validating the system on a mobile robot (TIAGo) to build maps from real-world sensor data in large-scale environments.
    \end{itemize}
\end{itemize}

\section{Related Work}
\label{sec:related_work}

Our work builds upon \glspl{3dssg} to bridge the gap between structured environment representations and open-set perception, integrating the knowledge-level expectation strategies of ExPrIS \citep{renz2026expris} with the language-driven exploration capabilities of LIEREx \citep{igelbrink2026lierex}. In the following, we review relevant approaches in \glspl{3dssg} and \gls{vlfm} integration for semantic mapping.

\subsection{3D Semantic Scene Graphs}

A \gls{3dssg} serves as a structured abstraction of the environment. As identified in our recent survey on online knowledge integration \citep{igelbrink2024online}, this paradigm is uniquely positioned to bridge the gap between sub-symbolic sensor data and high-level symbolic reasoning \citep{Armeni2019-ip, Rosinol2020-wi}. 
In this representation, nodes correspond to physical entities (ranging from individual objects to rooms and buildings) while edges encode the spatial or logical relationships between them \citep{Wald2020-yj, Hughes2022-fg}. 
By organizing these entities, often in a hierarchical manner, \glspl{3dssg} provide a flexible backend for complex downstream tasks such as navigation, task planning, and manipulation \citep{Kim2020-we, Agia2022-mi}. 
Furthermore, this structured format acts as a vital interface for external reasoning tools, enabling the integration of knowledge graphs and \glspl{llm} to perform open-set queries and common-sense reasoning directly on the mapped environment \citep{Lv2024-pn, Strader2023-je, ren2024explore}.

\subsubsection{Construction-Based Methods}
Pioneering works, such as \citet{Armeni2019-ip}, introduced hierarchical scene graphs to organize entities into layers (\eg buildings, rooms, objects). This concept was extended by \citet{Rosinol2020-wi} into 3D Dynamic Scene Graphs (DSGs), which include dynamic agents and capture actionable spatial information. The \textit{Hydra} framework \citep{Hughes2022-fg} further advanced this by enabling the real-time, incremental construction of DSGs from sensor data using spatial topology and loop closures. While these methods produce spatially consistent maps suitable for navigation, they typically rely on rigid heuristics and fixed ontologies, limiting their flexibility in handling semantic relationships beyond simple containment or adjacency.

\subsubsection{Prediction-Based Methods}
Conversely, learning-based approaches focus on inferring semantic relationships directly from data. \Citet{Wald2020-yj} introduced the 3DSSG dataset as the first benchmark for the prediction of \glspl{3dssg} from sensor data, proposing \glspl{gnn} as the main model architecture. To address the limitations of offline processing, \citet{wu2021scenegraphfusion} proposed \textit{SceneGraphFusion}, an incremental pipeline built on top of the 3DSSG dataset that fuses frame-wise graph predictions into a global graph representation. 
However, these purely data-driven methods often operate in a bottom-up fashion, failing to exploit the top-down knowledge priors (\eg structural expectations) that are central to hybrid systems. Furthermore, they rely heavily on labeled datasets and are difficult to extend to real domains. 
Our proposed architecture acts as a hybrid solution, utilizing the incremental update capabilities of prediction networks while maintaining a foundational backend that supports hierarchical grounding.

\subsection{Open-Set Semantic Mapping}
\label{subsec:related_openset}

The advent of \glspl{vlfm}, such as \gls{clip}~\citep{radford2021learning}, has shifted semantic mapping from closed-set vocabularies to open-set embedding spaces.

Early integrations of VLFMs into 3D mapping utilized implicit representations, such as Neural Radiance Fields (NeRFs). LERF \citep{kerrLERFLanguageEmbedded2023} and CLIP-Fields \citep{shafiullah2022clip} optimize scene-specific fields to query geometry with natural language. While visually accurate, these methods are computationally intensive and difficult to update. Geometry-based approaches like OpenScene \citep{pengOpenScene3DScene2023a} and ConceptFusion \citep{jatavallabhula2023conceptfusion} project dense pixel-aligned features onto point clouds or meshes. Although scalable, these representations often lack object-level abstraction, treating the environment as a collection of points rather than distinct entities, thus limiting scalability.

Recent works have sought to combine the structure of scene graphs with the flexibility of open-set embeddings. \Citet{werby2024hierarchical} proposed Hierarchical Open-Vocabulary Scene Graphs (HOV-SG), which aggregate \gls{clip} features within a hierarchical \gls{3dssg} inspired by Hydra to support language-based navigation. Similarly, ConceptGraphs~\citep{Gu2024-bq} and CLIO~\citep{maggio2024clio} construct object-centric maps that enable complex querying and task-dependent clustering. Our approach aligns with this trend but emphasizes the \gls{3dssg} as the \textit{foundational backend} for the entire mapping process, integrating prediction, construction, and open-set features into a single, unified consistency layer rather than treating them as separate downstream tasks.

\section{Mapping Approach}
\label{sec:mapping-approach}

The overarching design objective of our architecture is to establish the \gls{3dssg} as the central data structure and single source of truth for the entire mapping life cycle. 
Unlike conventional approaches that generate scene graphs as a post-processing step after the semantic mapping itself has finished, our framework aims to maintain the graph as the primary backend and enhance it continuously over time. 
This ensures that the graph encapsulates all relevant data regarding the current map state and remains topologically and semantically valid at each discrete time step. 
The system is designed for the incremental, online integration of high-bandwidth sensor data, leveraging a fully GPU-accelerated pipeline to guarantee computational efficiency suitable for real-world robotic applications.%

\subsection{Geometric Mapping}

Since our research focuses primarily on map representation, we deliberately decided to prioritize the development of our mapping architecture rather than introducing the complexity of a full SLAM framework.
Consequently, for real-world applications, our system relies on robust external pose estimates, \eg from an external tracking system or by localization in an existing map.
Leveraging available high-resolution polyhedral maps of our test environment, we employ MICP-L \citep{mock2024micp} for robot pose tracking using an Ouster OS0-128 LiDAR. Additionally, we align an RGB-D camera extrinsically to the LiDAR via ICP registration. We then utilize MICP-L to update the robot's pose estimation at the LiDAR timestamp nearest to each captured camera frame.
To enhance pose accuracy and mitigate extrinsic calibration and synchronization errors, we perform a subsequent offline pose-graph optimization step \citep{choi2015robust} directly on the camera poses.
Given that the initial MICP-L estimates exhibit minimal error and negligible drift, loop closures are detected efficiently via frustum checks.

\subsection{Scene Graph Backend}

Our approach adopts an instance-based, object-centric mapping paradigm, drawing inspiration from state-of-the-art dense semantic mapping methods such as ConceptGraphs~\citep{Gu2024-bq} and BBQ~\citep{linok2025beyond}, as well as the view-dependency logic of \citet{kassab2024bare}.

The backend data structure is implemented as a multi-layered \gls{3dssg}, organized into three distinct abstraction layers:
\begin{enumerate}
    \item A \textit{Frames Layer} that tracks 6D pose information and raw 2D segmentation masks for all keyframes processed;
    \item A \textit{Segments Layer} serving as an intermediate representation, tracking all currently active 3D object fragments including their raw geometry and extracted feature vectors. While currently integrated closely with the objects layer to reduce overhead, this layer is designed to support lazy, on-demand merging strategies in future iterations. Edges between the \textit{Segments} and \textit{Frames} layers naturally describe co-visibility relations and the stability of each segment; 
    \item An \textit{Objects Layer} that tracks persistent, consolidated object instances. This layer maintains the spatio-semantic relations between instances as well as their accumulated geometric and semantic feature information.
\end{enumerate}

\subsection{Mapping Pipeline}
\label{sec:pipeline}
The mapping pipeline is designed to be strictly incremental, ensuring the \gls{3dssg} backend transitions from one valid state to the next after each integration step.

\subsubsection{Segmentation and Feature Extraction}

For each incoming RGB-D frame, we first employ a pre-trained \textit{Segment Anything Model} (SAM) to extract a set of potentially overlapping segmentation masks. 
To ensure high-quality input, a filter bank immediately discards masks with low confidence, extreme aspect ratios, or insufficient size. 
Mask boundaries are further refined using depth discontinuities to prevent ``bleeding'' of segments across physical boundaries. The pipeline is agnostic to the specific SAM variant; in our experiments, we utilize FastSAM \citep{zhao2023fast} due to its favorable inference speed, which is critical for online operation.

Simultaneously, we extract dense visual feature vectors from the RGB image using the DINOv2 model \citep{oquab2023dinov2}. Inspired by the efficiency strategies in \citet{linok2025beyond} and MaskCLIP \citep{dong2023maskclip}, we utilize the local per-patch features directly from the model's intermediate layers rather than computing aggregated features for separate object crops. This design allows us to execute the heavy feature encoder only once per frame, significantly increasing throughput. Since these features are primarily used for short-term data association between frames, the lack of global context in patch features is acceptable at this stage. Per-segment feature descriptors are obtained by aggregating the patch features within the corresponding refined segmentation masks.

\subsubsection{Local Graph Generation}
The depth data is projected into a 3D point cloud for each valid segment. To mitigate sensor noise and remove segmentation artifacts (\eg flying pixels), we filter these points using a custom CUDA-based implementation of the DBSCAN algorithm. This implementation utilizes a parallel, batched union-find approach to maximize performance on the GPU. The resulting point clouds are voxelized to ensure uniform point density across the map.

These processed 3D segments are initially instantiated in a \textit{local} \gls{3dssg}. This local graph mirrors the structure of the global graph but describes only the information contained in the current frame. This intermediate representation provides a crucial hook for additional graph-based refinement and prediction modules (see Section~\ref{sec:ipp}) before the data is committed to the global state.

\subsubsection{Data Association and Global Integration}
The integration of the \textit{local} graph into the persistent \textit{global} graph is performed via a robust two-stage matching process.

\paragraph{Stage 1: Greedy Association}
First, we employ a conservative greedy matching strategy. We compute an affinity matrix based on the Intersection over Union (IoU) of 3D bounding boxes and the cosine similarity of the DINOv2 feature vectors. Any \textit{local} segment that overlaps with a \textit{global} segment and exceeds a strict similarity threshold is merged immediately. Unmatched segments are instantiated as new nodes in the global graph. This stage is designed to be safe, merging only unambiguous matches to prevent corruption of the map. This simple matching does not perform any conflict resolution and does not throw away potentially valuable segments from the segmentation stage. 

\paragraph{Stage 2: Active Refinement}
Greedy merging alone often leads to over-segmentation and the accumulation of spurious artifacts due to temporal inconsistencies in the SAM predictions. Previous works typically addressed this via periodic, computationally expensive offline passes over the entire map \citep{Gu2024-bq}. To maintain the \gls{3dssg} as a real-time ``best estimate,'' we introduce an active refinement stage that runs immediately after greedy merging.
Crucially, this refinement considers only the \textit{active subset} of the graph---nodes that were either modified or newly created in the current step. These nodes are matched against their spatial neighbors using voxel-grid overlap. For each node in this subset, the system evaluates potential merge candidates based on feature stability and overlap. This process naturally resolves conflicts and ``daisy-chain'' merge effects, effectively fusing fragmented raw segments into coherent object instances that approach the semantic granularity of the underlying foundation model. In future work, we aim to augment this geometric heuristic with a learned semantic edge prediction model to infer merges based on learned spatial compatibility.

\subsection{Vision-Language Feature Integration}

To facilitate open-vocabulary queries, we integrate additional \gls{clip} features directly into the graph nodes. Similar to our DINOv2 integration above, we circumvent the computational bottleneck of running the \gls{clip} encoder on individual mask crops. Instead, we adopt the MaskCLIP \citep{dong2023maskclip} paradigm: we extract per-patch features from the penultimate transformer layer of a standard CLIP model and post-process them using the model's native attention and projection weights. 

While efficient, this method yields features that are strongly biased towards local textures rather than global object semantics (\eg encoding ``fabric pattern'' rather than ``couch''). To compensate, we implement a \textit{gated integration} mechanism. We compute an additional global CLIP embedding for the entire frame and modulate the local patch features based on their cosine similarity to this global context. This enhances the semantic richness of segments that align with the scene context while allowing distinct outliers to retain their specific characteristics.

Finally, we compute a \textit{view quality score} for each segment integration, derived from geometric factors such as the segment's relative size and distance from the image center \citep{kassab2024bare}. This score acts as a confidence weight during feature aggregation: features extracted from optimal viewpoints contribute significantly to the node's persistent embedding, while oblique or occluded observations have minimal impact. This ensures that the open-set representation remains robust against noisy fleeting observations.

\section{Incremental Predicate Prediction}
\label{sec:ipp}

The presented mapping framework, as well as existing approaches like ConceptGraphs~\citep{Gu2024-bq} and CLIO~\citep{maggio2024clio}, show great flexibility towards environments with complex objects and structures. However, edges in these scene graphs usually have no deeper semantic meaning and are only included for hierarchy or spatial structure.
Relationships such as \emph{on top of} or \emph{attached to} are useful information for reasoning algorithms such as task planners, or even \glspl{llm} used instead of a planner. For this reason, we are combining the mapping pipeline with methods from scene graph prediction to predict semantic labels for relationships between objects. By introducing semantic relationships between open-set objects, we maintain the flexibility of open-set scene graphs while enhancing the semantic part of the \gls{3dssg}.

Using a closed-set prediction approach is less flexible than using an \gls{llm}, VLM, or \gls{vlfm}. However, short inference times are a requirement for real-time mapping; therefore, we aim for a smaller architecture that could even be deployed on the edge.

\subsection{Prediction Graph Model}

\begin{figure}[tb]
    \centering
    \includegraphics[width=\linewidth]{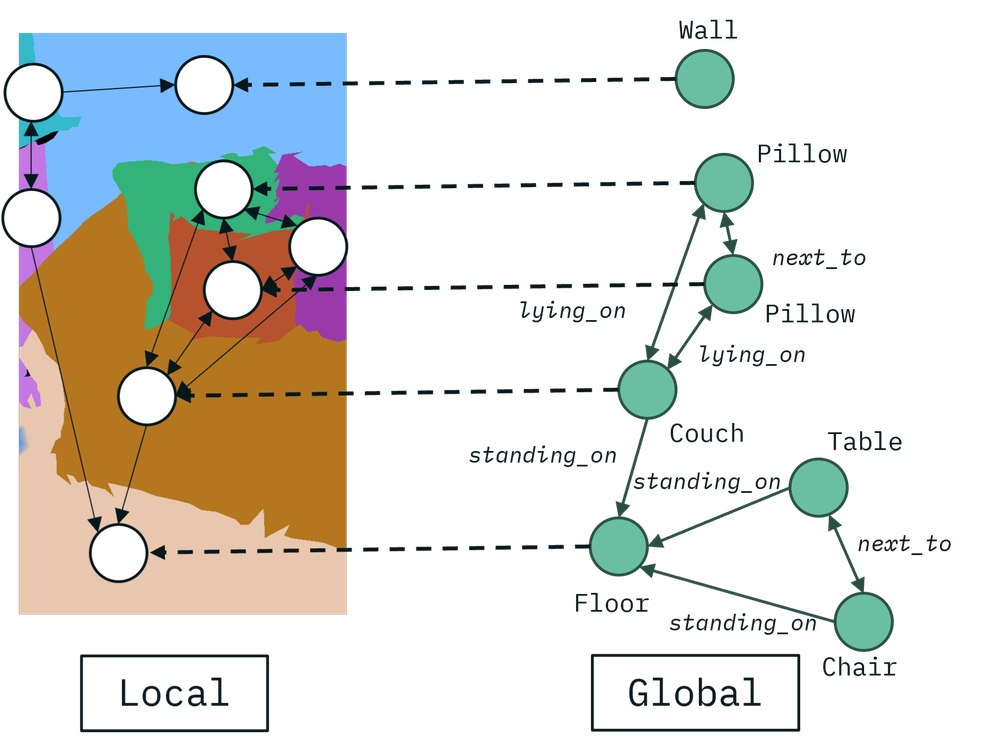}
    \caption{The heterogeneous graph model for Incremental Predicate Prediction (IPP) from \citet{Renz2025-uk}. The local graph corresponds to the segments layer or the frame-only local scene graph, the global graph corresponds to the objects layer or the fully integrated and refined graph from the mapping pipeline described in Section~\ref{sec:mapping-approach}. Dashed edges are between matched objects between the local frame and global map to provide information flow during message passing. While object classes are displayed here for clarity, the actual global scene graph only contains open-set features.}
    \label{fig:hetero-graph}
\end{figure}

To infer the semantic relationships between objects in the graph, we base the prediction on the heterogeneous incremental graph model introduced by \citet{Renz2025-uk}. The model consists of a local/global structure: a global layer, which contains the mapped scene graph from previous time steps, and a local layer, which contains the scene graph built only on the local sensor data frame. The overall goal for each time step is to predict the predicate labels between objects in the local layer while keeping the open-set features, in contrast to common scene graph prediction approaches like \citet{Wald2020-yj} where the object classes are predicted as well. The layers in the graph prediction model correspond exactly to the local and global scene graphs introduced in Section~\ref{sec:pipeline}. Predicate predictions from the local layer are incrementally integrated into the global scene graph. Since segments in the local frame have (for the most part) a corresponding object in the global graph, we can connect them in the graph model for prediction (see Figure~\ref{fig:hetero-graph}). This way, information from the previously predicted semantic predicates can flow into the local prediction during the message-passing process, assisting predictions of predicates that have been observed before.

\textit{Node features:} For each node in the local and the global scene graph, we sample 256 voxel points. Additionally, we construct a geometric descriptor of each node's 3D segment in the map/frame, analogous to SceneGraphFusion~\citep{wu2021scenegraphfusion} and the approach by \citet{Renz2025-uk}: 
\[
[c, \text{std}, l, w, h, L, V]
\]
where $c$ is the 3D segment's oriented bounding box center, $\text{std}$ the standard deviation of each point from the center, $l$, $w$, and $h$ the bounding box dimensions, $L$ the bounding box's biggest side, and $V$ the bounding box volume. Furthermore, we use the DINOv2 features the mapping framework provides for each segment.

\textit{Edges:} A standard practice to determine the presence of an edge between two objects is by connecting objects within a 0.5~m radius to each other. However, we find that this leads to overly dense graphs and does not represent a scene's structure sufficiently. Therefore, we decided to connect only touching objects and predict the labels of those edges. To find touching object pairs, we add 5~cm padding to each segment's bounding box and check for intersections. To connect matching nodes between the local and global layers, we reuse the pairs found by the greedy association step in Section~\ref{sec:pipeline} and add unlabeled directed edges between matching nodes between the local and global layers.

\subsection{Graph Neural Network}

\begin{figure}[tb]
    \centering
    \includegraphics[width=1.0\linewidth]{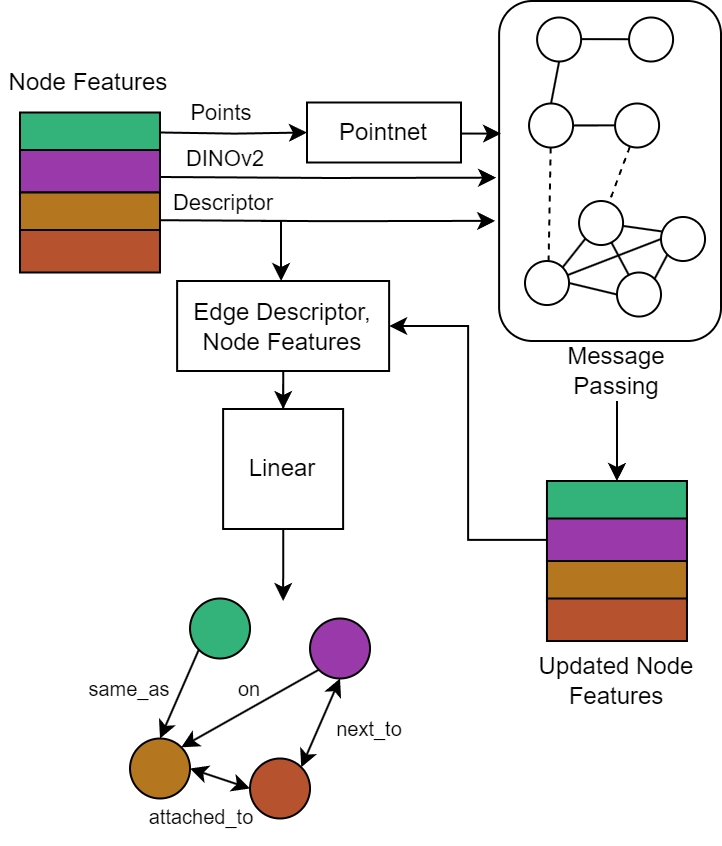}
    \caption{Graph neural network architecture for Incremental Predicate Prediction (IPP).}
    \label{fig:gnn}
\end{figure}

\begin{figure*}[tb]
\centering
\begin{subfigure}[c]{0.32\textwidth}
    \centering
    \includegraphics[width=\textwidth]{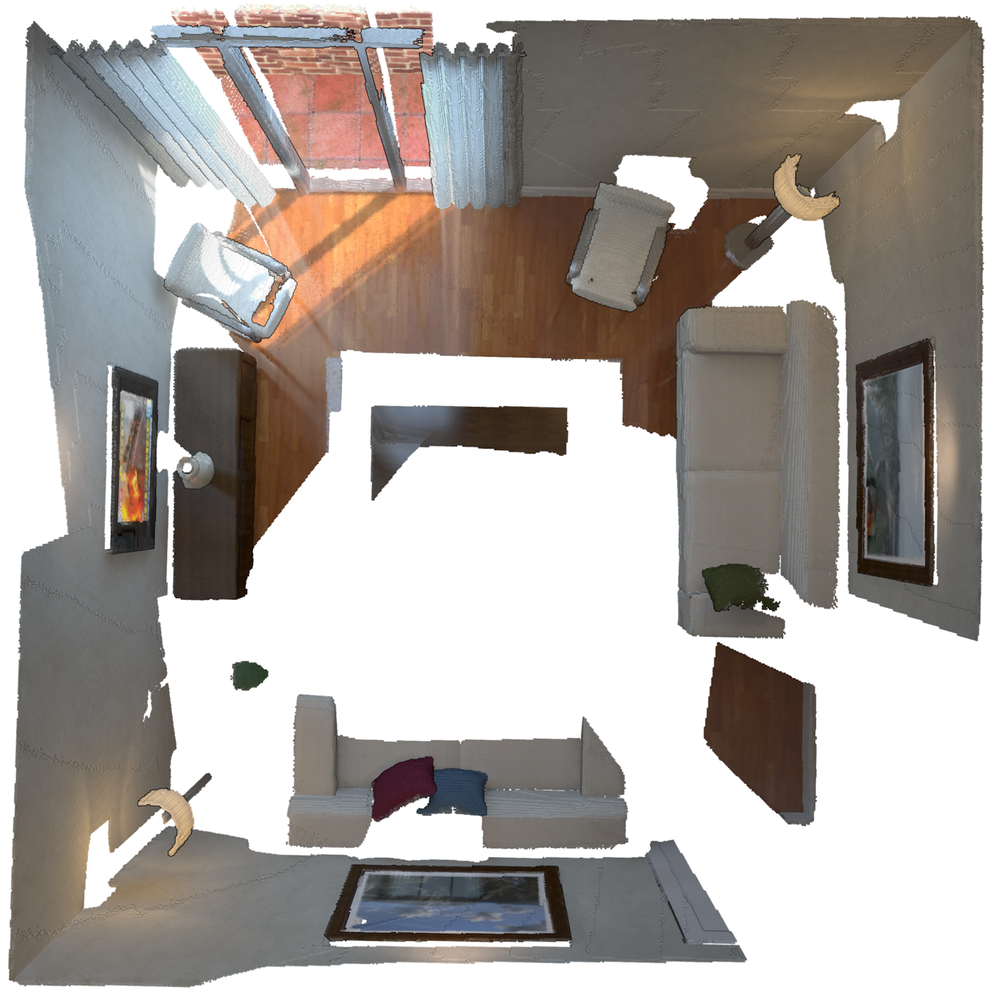}
    \subcaption[]{RGB}
\end{subfigure}
\begin{subfigure}[c]{0.32\textwidth}
    \centering
    \includegraphics[width=\textwidth]{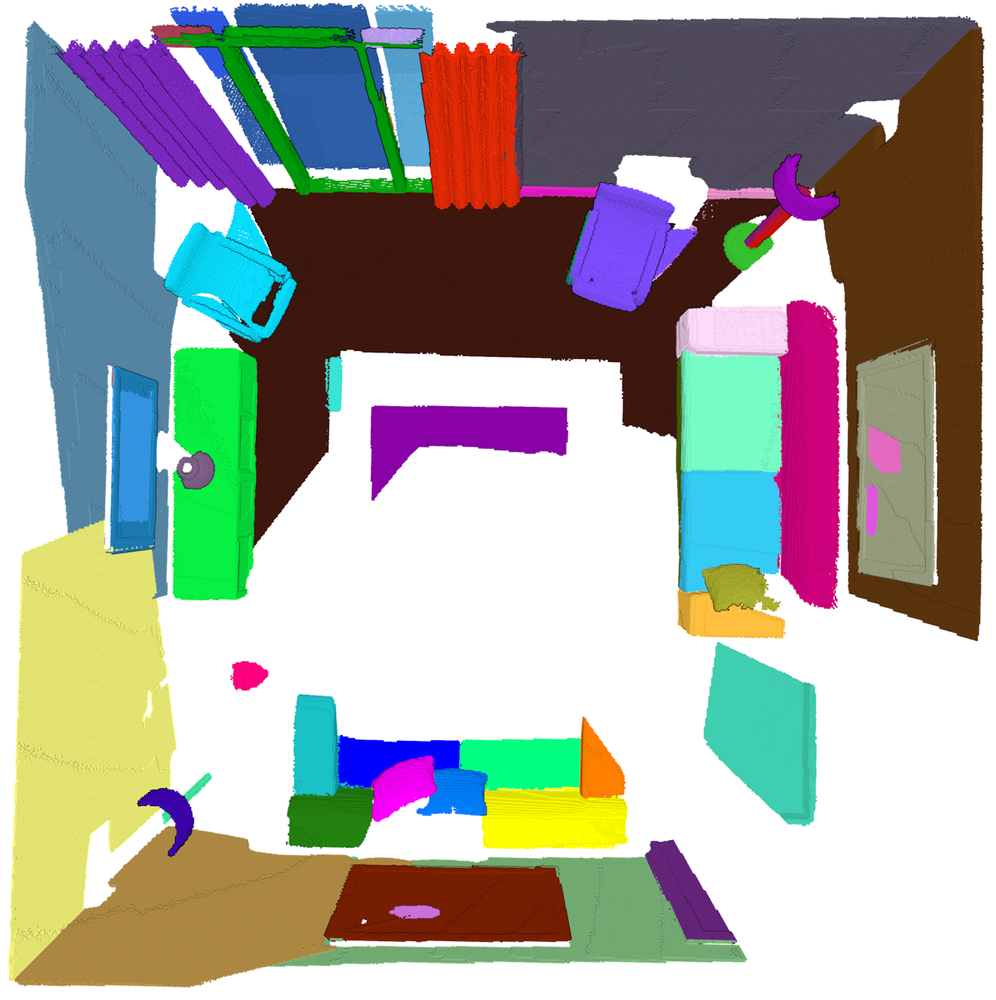}
    \subcaption[]{Instances}
\end{subfigure}
\begin{subfigure}[c]{0.32\textwidth}
    \centering
    \includegraphics[width=\textwidth]{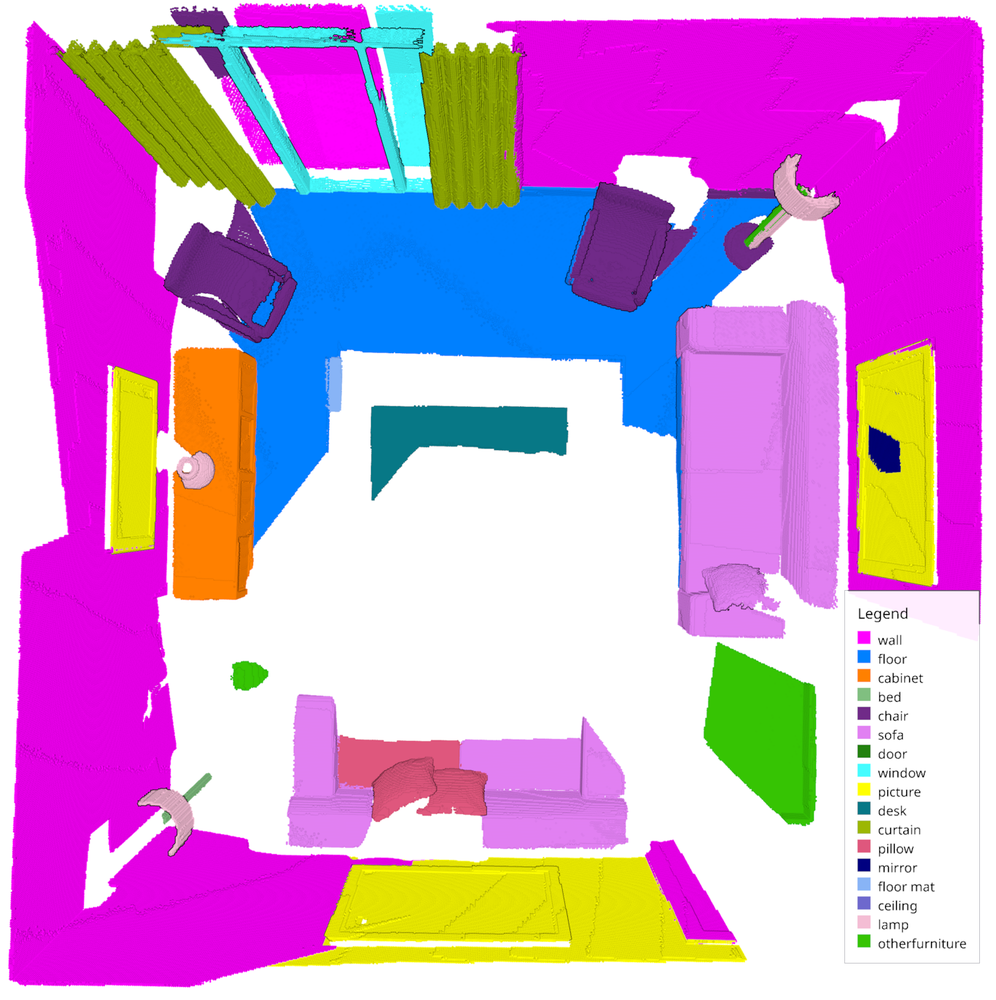}
    \subcaption[]{Segmentation}
\end{subfigure}
\caption{Preliminary results of the mapping pipeline on the small-scale ICL dataset. The segmentation was obtained by comparing each object's CLIP feature with the embeddings of the NYU-40 label set.}
\label{fig:semantic-map-results}
\end{figure*}

After constructing the prediction graph from the mapping pipeline's scene graph, a heterogeneous graph neural network is used to classify the edge labels.%
\footnote{The implementation of the Incremental Predicate Prediction network described in this subsection is ongoing work as well as the tight integration with the real-time mapping pipeline. The following describes the architecture of this component.}
Descriptor, DINO features, and points are used as node features for the local and the global layers, although the points are embedded into a latent vector using PointNet~\citep{Qi2017-wl} first. Message passing is done using a heterogeneous GraphSage~\citep{Hamilton2017-nc}, where edges in the local and global layers respectively, as well as the intra-layer edges, are modeled using separate linear layers.

While both node types provide the same kind of information, the heterogeneity comes from the already predicted, integrated edges in the global scene graph, which serves as a semantic prior and is included in message passing. For this, we extend the GraphSage message passing for the global layer to include a text embedding such as GloVe \citep{pennington2014glove} and a respective linear layer:

\begin{equation}
    x_i' = \gamma\Bigl(x_i,  \Theta \{  x_j + \phi(e_{ji}) \forall j \in \mathcal{N}(i)\} \Bigr)
\end{equation}

where $\gamma$ is a feed-forward layer with non-linear activation, $\Theta$ is an aggregation function like mean or sum, $\phi$ is a linear layer, and $x_i$, $x_j$, and $e_{ij}$ are target node-, neighbor node-, and edge features respectively.

After two layers of message passing, the final edge features are created by concatenating the two updated node features of the nodes connected by the respective edge, $n_i$ and $n_j$, as well as a combined edge descriptor: 
\[
[c_j - c_i, \; \text{std}_j - \text{std}_i, \; \log(\tfrac{l_j}{l_i}, \tfrac{w_j}{w_i}, \tfrac{h_j}{h_i}), \; \log(\tfrac{L_j}{L_i}), \; \log(\tfrac{V_j}{V_i})]
\]

Concatenated features and descriptors are then passed through a feed-forward network to obtain the final classification logits (see Figure \ref{fig:gnn}).

The predicate prediction is modeled as a multilabel problem, since this way edges can have no labels and therefore would not be integrated into the global graph.

\subsection{Training Data Preparation}

We construct the training data for the Incremental Predicate Prediction using 3D scenes from the 3RScan dataset \citep{Wald2019-ki} as well as the predicate labels from the 3DSSG dataset \citep{Wald2020-yj}. 

We incrementally build a 3D map using the mapping pipeline while storing both the local scene graph and the global scene graph at that timestep as described in Section~\ref{sec:ipp}, with sampled points, DINO features, and descriptors, as well as edges between touching objects, classified global edges, and matched nodes connected by intra-layer edges.

\section{Preliminary Results}
\label{sec:results}

\subsection{Mapping Results}

To validate the fundamental capabilities of our proposed architecture, we conducted preliminary experiments on the ICL RGB-D dataset \citep{handa:etal:ICRA2014}. 

Figure~\ref{fig:semantic-map-results} presents the qualitative results on the \texttt{kt1} trajectory, displaying the reconstructed color mesh, the instance segmentation, and the resulting semantic segmentation for a small scale environment. 
The semantic labels were generated by querying the open-set CLIP features of each node against the text embeddings of the NYU-40 label set. 

As observed in the instance segmentation layer, the system successfully isolates major structural elements. 
However, certain areas exhibit over-segmentation artifacts. 
These occur primarily where the active refinement stage (see Section~\ref{sec:pipeline}) has not yet accumulated sufficient evidence, such as consistent spatial overlap or feature similarity, to confidently merge adjacent segments.
This behavior is a consequence of our conservative merging strategy, designed to prevent the irreversible fusion of distinct objects.

Regarding semantic accuracy, the integration of CLIP features proves effective for distinct, well-separated objects within uncluttered scenes. 
However, the performance degrades in highly cluttered areas or with many very small objects. 
This limitation stems from the current inherent resolution constraints of the used MaskCLIP approach, where the patch size of the underlying vision transformer limits the granularity of the extracted features. 
Addressing this resolution mismatch by improving the feature aggregation remains a key objective for our future work.

\subsection{Predicate Prediction Integration}

\begin{figure}[tb]
    \centering
    \includegraphics[width=1.0\linewidth]{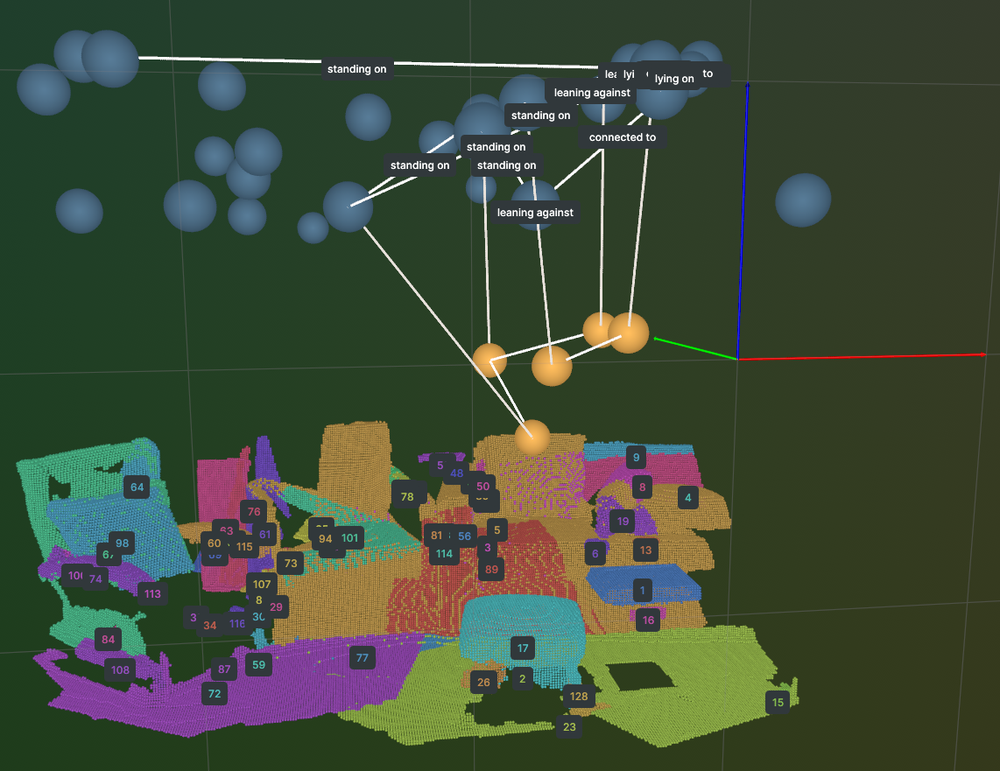}
    \caption{Visualization of local scene graph (yellow) and global scene graph (blue) with inter- and intra-layer edges with a mapped scene from the 3RScan dataset. Predicted edges are integrated into the global graph using ground truth annotations from \citet{Wald2020-yj}.}
    \label{fig:PoC-ipp}
\end{figure}

\begin{figure*}[ht]
\centering
\begin{subfigure}[c]{0.33\textwidth}
    \centering
    \includegraphics[width=\textwidth]{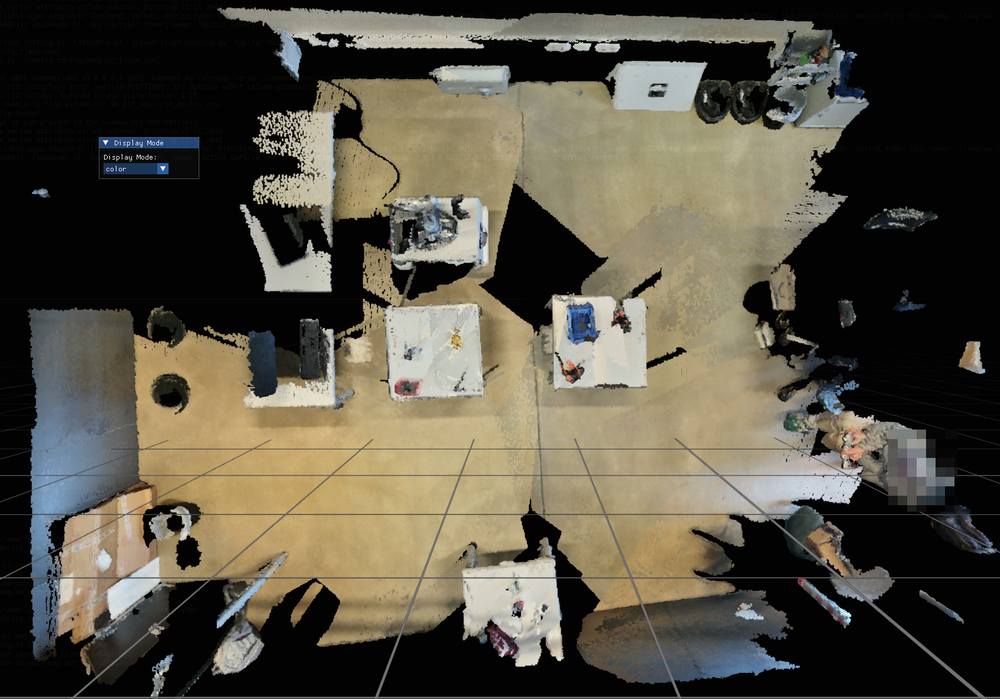}
    \subcaption[]{RGB Reconstruction}
    \label{fig:semantic-map-real_rgb}
\end{subfigure}
\begin{subfigure}[c]{0.33\textwidth}
    \centering
    \includegraphics[width=\textwidth]{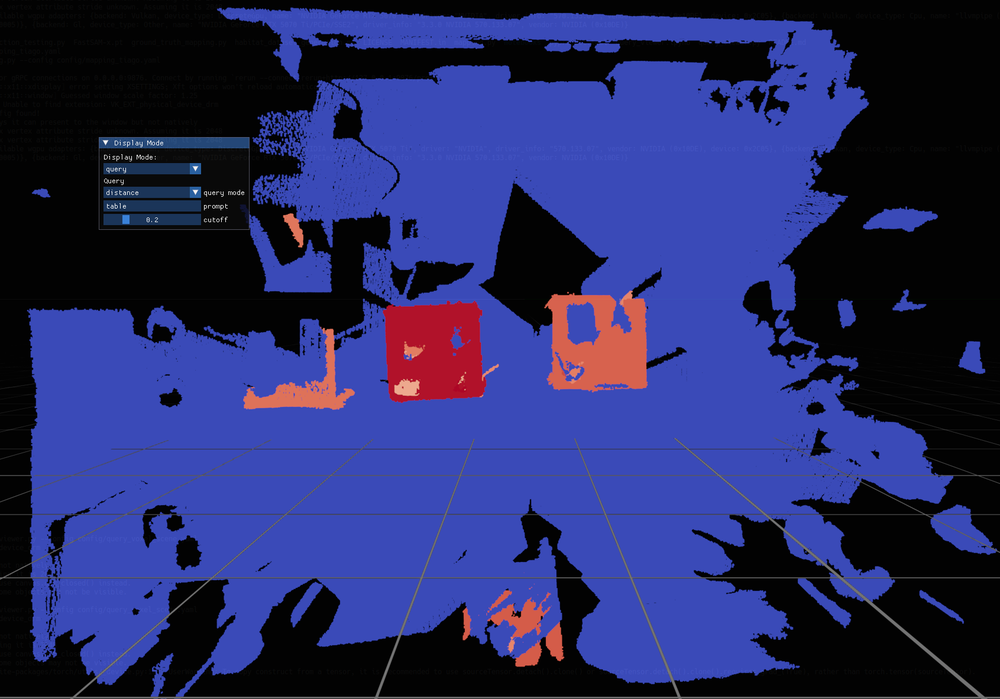}
    \subcaption[]{Query: ``Table''}
\end{subfigure}
\begin{subfigure}[c]{0.33\textwidth}
    \centering
    \includegraphics[width=\textwidth]{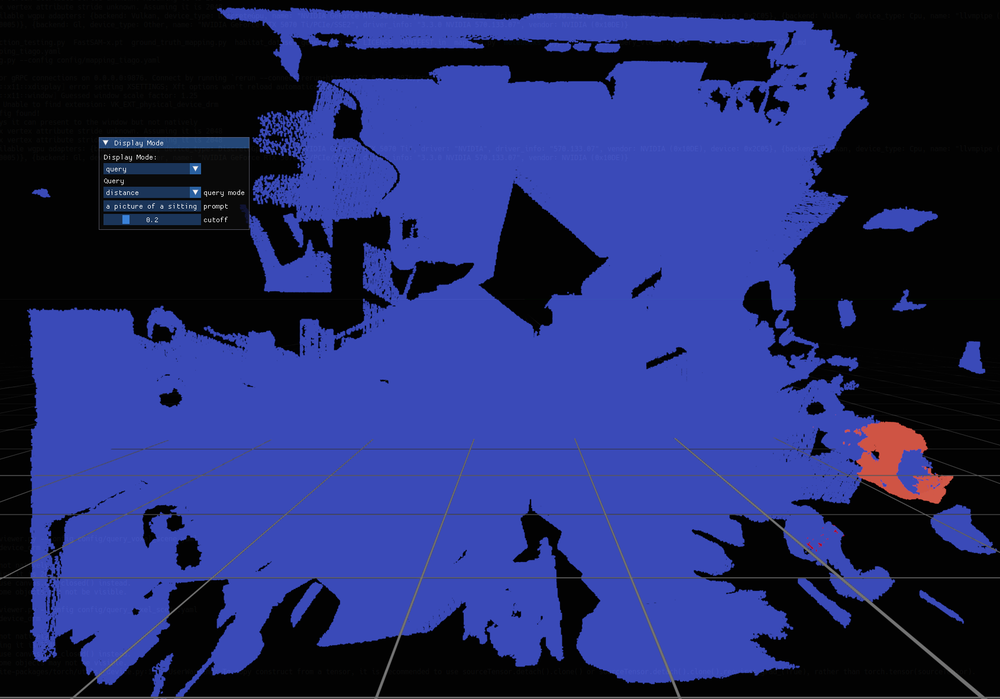}
    \subcaption[]{Query: ``A photo of a sitting man''}
    \label{fig:semantic-map-real_d}
\end{subfigure}
\caption{Mapping results and sample queries on real-world data. (a) Successful map integration given the estimated poses. (b) \& (c) Visualization of cosine similarity between instances and the respective text queries.}
\label{fig:semantic-map-real}
\end{figure*}

\begin{figure*}[ht]
\centering
\begin{subfigure}[c]{0.495\textwidth}
    \centering
    \includegraphics[width=\textwidth]{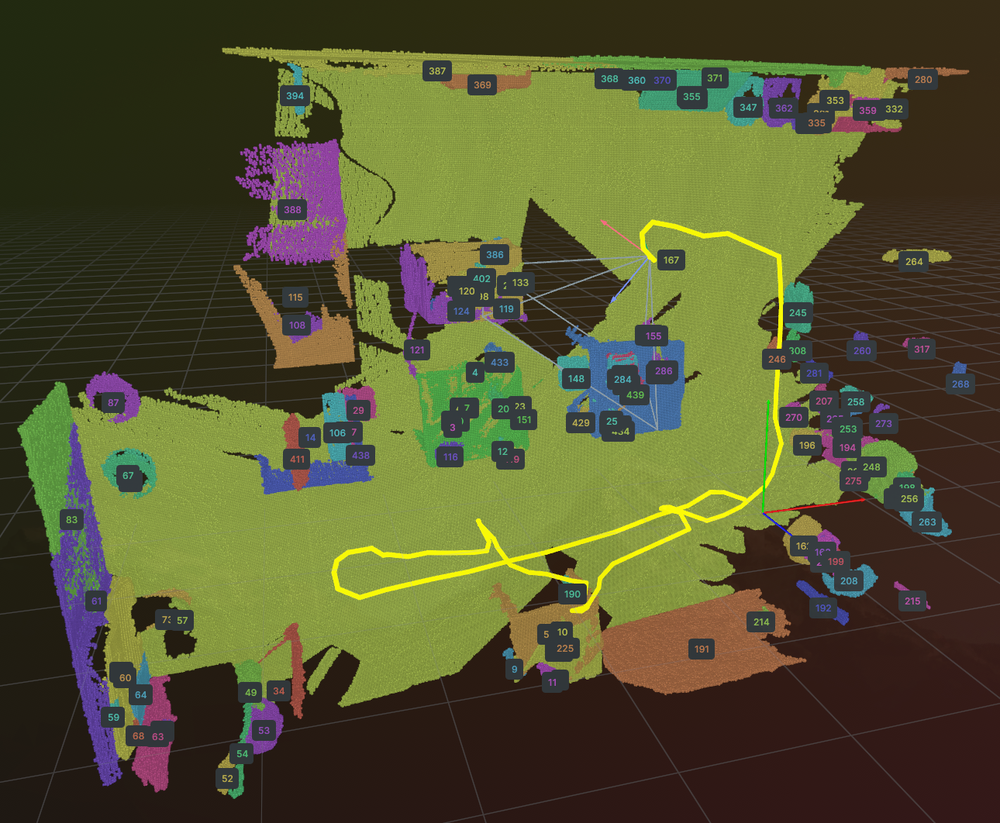}
    \subcaption[]{Instance segmentation on unfiltered depth data}
\end{subfigure}
\begin{subfigure}[c]{0.495\textwidth}
    \centering
    \includegraphics[width=\textwidth]{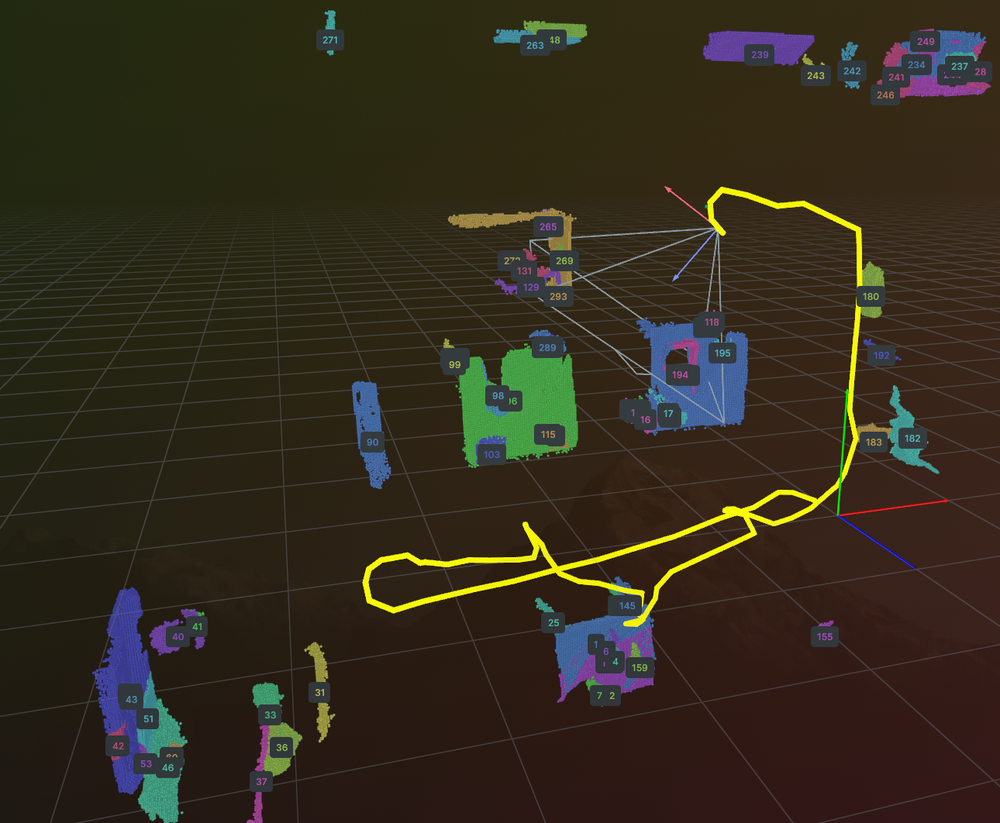}
    \subcaption[]{Instance segmentation on filtered data (simulated panoptic)}
\end{subfigure}
\caption{Comparison of mapping results with and without geometric filtering. (a) Mapping performed on raw unfiltered depth data. (b) Points corresponding to the static building map (used for localization) are filtered out, effectively mimicking panoptic segmentation to remove structural \emph{stuff} classes.}
\label{fig:semantic-map-filtered}
\end{figure*}

To showcase the Incremental Predicate Prediction (IPP), we display the integration of the predicted predicates into the 3DSSG on example scenes from the 3RScan dataset (see Figure~\ref{fig:PoC-ipp}). RGB-D frames are rendered from the 3D scene reconstructions. For each frame, the predicate predictions are merged into the global scene graph and used in the next time step. Edges without predictions are not integrated. 

Training, inference and actual integration using the described \gls{gnn} is part of future work.

\subsection{Real-World Application}

To demonstrate real-world applicability of our proposed framework, we evaluated our system using the TIAGo mobile robotic platform.
Initial pose estimates are provided by MICP-L~\citep{mock2024micp} using an Ouster OS0-128 LiDAR against a high-resolution triangle mesh of the environment\footnote{The model can be found at: \url{https://github.com/DFKI-NI/pbr_tools/tree/jazzy/pbr_gazebo/models/cic/meshes}}, which includes only the static building structure.
RGB-D sequences were recorded at $1280\times720$ resolution via a Femto Bolt ToF RGB-D camera. 
We processed a subset of 440 frames from the resulting frame-pose pairs in order to generate the map.
The resulting map and the selected sample queries are presented in Figures~\ref{fig:semantic-map-real} and \ref{fig:semantic-map-filtered}. While instances are not as cleanly segmented as in the synthetic ICL dataset (\cf Figures~\ref{fig:semantic-map-results} and \ref{fig:semantic-map-filtered}), the system successfully integrates the majority of instances into a coherent map. Most errors can be attributed to sensor noise, minor pose inaccuracies, and FastSAM over-segmentation, which is particularly noticeable on objects that were only partially observed.

Furthermore, we conducted an experiment using Rmagine \citep{mock2023rmagine} to filter the input data. By matching depth data against the building model used for localization, we mimic a panoptic segmentation step to filter out \emph{stuff} classes. 
Figure~\ref{fig:semantic-map-filtered} illustrates the impact of this filter: floors and walls are effectively removed while most \emph{things} remain, significantly reducing the number of spurious instances in the final map.

\section{Discussion and Future Outlook}
\label{sec:discussion}

In this paper, we presented a hybrid mapping architecture that establishes the \gls{3dssg} not as a derivative output, but as the foundational backend for robot perception. By moving away from monolithic geometric maps and towards an explicit semantic knowledge structure, we have demonstrated a powerful and flexible framework that bridges the gap between sub-symbolic sensory data and symbolic reasoning -- a core objective of hybrid intelligence.

A key strength of this architecture is its representation-agnostic nature. While our current implementation utilizes a voxel-based representation to efficiently handle the scalability requirements of large-scale environments, the high-level symbolic layer is designed to be independent of the specific underlying geometric format. Consequently, the backend could readily support \glspl{tsdf}, meshes, or neural fields as the geometric source, provided they can be grounded into the graph's node structure. This modularity not only offers significant flexibility for different robotic platforms but also ensures that the system maintains a stable, verifiable structure that aligns with human concepts. As argued in Section~\ref{sec:introduction}, this explicit structure is what enables the seamless integration of top-down expectations and open-vocabulary queries, capabilities that are difficult to achieve with purely implicit or geometric representations.

Despite these advantages, the current implementation of the Incremental Predicate Prediction (IPP) module faces a limitation common to many neuro-symbolic approaches: a reliance on supervised learning. The graph neural networks employed for relation inference currently require extensive ground-truth labeling, such as that provided by the 3DSSG dataset~\citep{Wald2020-yj}. Hence, such a dependence on annotated datasets currently restricts the system's ability to generalize to novel environments where such dense relational labels are unavailable.

To overcome the bottleneck of supervised relational learning, our future work will focus on inferring spatial relations in an unsupervised manner. We aim to adopt strategies similar to the SEMAP framework~\citep{deeken2015semap}, which extracts geometric spatial relations (e.g., \emph{on}, \emph{close to}) directly from the map topology without requiring training data. We are currently developing a unified approach for object detection, 6D pose estimation, and spatial relation inference that leverages these geometric heuristics \citep{niecksch2024mesh}. By replacing the supervised prediction head with these unsupervised geometric extractors, we aim to enable the autonomous construction of rich semantic scene graphs in entirely open and unlabelled worlds.

\section{Acknowledgements}
This work is supported by the ExPrIS and LIEREx projects through grants from the German Federal Ministry of Research, Technology and Space (BMFTR) with grant numbers 16IW23001 and 16IW24004.
The DFKI Niedersachsen (DFKI NI) is sponsored by the Ministry of Science and Culture of Lower Saxony and the VolkswagenStiftung.

\bibliography{bibliography}

\end{document}